\documentclass[letterpaper]{article} % DO NOT CHANGE THIS
\usepackage[preprint]{preprint_style}
\usepackage[hyphens]{url}  % DO NOT CHANGE THIS
\usepackage{graphicx} % DO NOT CHANGE THIS
\usepackage{natbib}  % DO NOT CHANGE THIS AND DO NOT ADD ANY OPTIONS TO IT
\usepackage{caption} % DO NOT CHANGE THIS AND DO NOT ADD ANY OPTIONS TO IT
\usepackage{algorithm}
\usepackage{algorithmic}

\usepackage[table]{xcolor} % \textcolor, \rowcolor, \definecolor
\usepackage{amssymb}       % \checkmark
\usepackage{amsmath}       % align, \operatorname, \text
\usepackage{makecell}      % \makecell
\usepackage{siunitx}       % 表格中的 S 列

\usepackage{booktabs} % 解决 \toprule, \midrule, \bottomrule 报错
\usepackage{multirow} % 解决 \multirow 报错

\usepackage{graphicx}
\usepackage{tabularx} % 必须引入这个宏包来实现自动换行

\usepackage{newfloat}
\usepackage{listings}
\DeclareCaptionStyle{ruled}{labelfont=normalfont,labelsep=colon,strut=off} % DO NOT CHANGE THIS
\floatstyle{ruled}
\newfloat{listing}{tb}{lst}{}
\floatname{listing}{Listing}

\usepackage{booktabs}

\title{Understand Before Detect: Vision--Language Learning for Omni-Domain Infrared Small Target Detection}
\author{
    Haoyang Yuan\textsuperscript{\rm 1},
    Boyang Li\textsuperscript{\rm 1},
    Yingqian Wang\textsuperscript{\rm 1},
    Yimian Dai\textsuperscript{\rm 2},
    Nuo Chen\textsuperscript{\rm 3},\\
    Xinfei Huang\textsuperscript{\rm 1},
    Shuqi Yi\textsuperscript{\rm 1},
    Zaiping Lin\textsuperscript{\rm 1},
    Weidong Sheng\textsuperscript{\rm 1},
    Wei An\textsuperscript{\rm 1}
}
\affiliations{}

\makeatletter
\let\arxivsavedmaketitle\maketitle
\let\arxivsavedatmaketitle\@maketitle
\makeatother

\begin{document}

\maketitle

\begingroup
\renewcommand{\thefootnote}{}
\makeatletter
\long\def\@makefntext#1{\noindent #1}
\makeatother
\footnotetext{\tiny\raggedright
\textsuperscript{\rm 1}National University of Defense Technology, Changsha, China;
\textsuperscript{\rm 2}College of Computer Science, Nankai University, Tianjin, China;
\textsuperscript{\rm 3}Peking University, Beijing, China.\\
\{yuanhaoyang25, liboyang20, wangyingqian16, huangxinfei25, yishuqi21, linzaiping, shengweidong, anwei\}@nudt.edu.cn\\
yimian.dai@gmail.com; chennuo@pku.edu.cn\\
The code, datasets, and models will be released at \pdfstartlink attr{/Border [0 0 0]} user{/Subtype /Link /A << /S /URI /URI (https://github.com/shushengyuan/OmniIRSTD-VL/tree/main) >>}\textcolor{blue}{this https URL}\pdfendlink.}
\endgroup

\begin{abstract}
Omni-domain infrared small target (IRST) detection is crucial for infrared surveillance, yet remains challenging due to heterogeneous imaging domains and inconsistent target characteristics.
Previous deep learning-based methods have been developed for visual-only paradigms and achieved promising performance on domain-specific tasks. However, existing methods follow the task-specific supervised learning paradigm. This paradigm simplifies the full-scene infrared observations to sparse target supervision, discarding the semantics that remain invariant across heterogeneous domains. Consequently, detection performance suffers substantially under domain shifts. To handle this issue, we introduce \textbf{``understand before detect''}, a paradigm that formulates omni-domain IRST detection as an understanding-driven process, where holistic infrared target understanding precedes precise detection. Building on this paradigm, we propose \textbf{JinSight}, which first develops holistic IRST understanding through language supervision and then transfers the learned cross-domain representations to precise small-target detection. By grounding infrared representations in language semantics, JinSight enables a single model to generalize across heterogeneous infrared domains. We then introduce Latent Semantic Interaction (LSI), which exchanges language-aligned global semantics with fine-grained spatial features in a compact low-rank space. To address the lack of multimodal omni-domain IRST benchmarks, we build \textbf{OmniIRST-VL}, the first large-scale, highly diverse vision--language dataset for omni-domain IRST detection. It comprises over 39k annotations across six complementary instruction tasks covering both scene-level understanding and target-centric reasoning.
Extensive experiments demonstrate that JinSight consistently outperforms existing state-of-the-art approaches and achieves over 14\% IoU improvement on the WideIRSTD benchmark, validating its effectiveness for omni-domain IRST detection.
\end{abstract}

\section{Introduction}

\label{sec:intro}
\begin{figure}[t]
	\centering
	\includegraphics[width=\columnwidth]{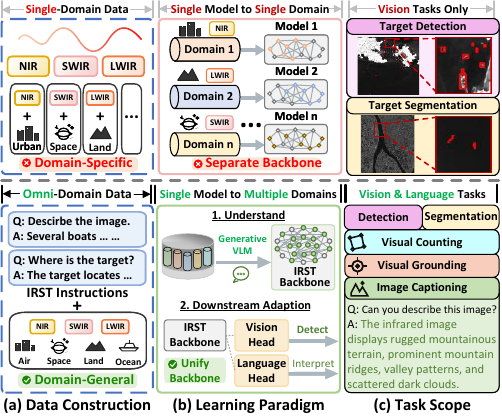}
	\caption{Comparison between conventional domain-specific IRST learning (top) and JinSight (bottom) in terms of (a) data construction, (b) learning paradigm, and (c) task scope. Following the ``understand before detect'' paradigm, JinSight first learns a unified IRST-aware backbone from omni-domain instructions through a generative VLM \cite{InternVL} and then adapts it to detection and understanding tasks.}\label{fig:Fig1}
\end{figure}

Infrared small target (IRST) detection has been widely used in applications such as traffic monitoring \cite{traffic}, autonomous driving \cite{autonomous_driving}, and anti-UAV systems \cite{antiuav}, owing to the passive, illumination-independent, and all-weather nature of infrared sensing. With growing demand for large-scale real-world deployment, IRST detection systems are deployed across highly heterogeneous settings: sensors mounted on spaceborne, airborne, and ground-based platforms cover different spectral bands (e.g., NIR, SWIR, LWIR) and observe scenes ranging from open skies to sea surfaces. This motivates omni-domain IRST (Omni-IRST) detection \cite{auxdet}, which seeks to detect targets across heterogeneous infrared domains with a unified detector (Figure~\ref{fig:Fig1}). However, variations across these domains induce dramatic shifts in scene context, background characteristics, and target appearance. Such shifts, compounded by the weak responses and sparse spatial footprints of infrared small targets, pose substantial challenges to learning shared representations that preserve target–background discriminability.

Existing IRST methods have mainly improved visual perception within specific domains. CNN-based models \cite{dnanet} exploit local inductive biases, multi-scale feature fusion, and dense cross-layer connections to preserve small target responses, while transformer-based or hybrid models \cite{sctransnet} introduce long-range contextual modeling to suppress background clutter. However, these methods still follow the conventional target detection paradigm, casting IRST detection as pixel-wise binary target--background prediction. While this binary abstraction facilitates the use of mature detection frameworks, it obscures the inherent information richness of omni-domain infrared data---namely, target–context relationships and scene-level semantic cues. This formulation encounters three critical limitations in Omni-IRST detection: (1) \textbf{Domain heterogeneity collapse:} diverse background structures, clutter patterns, and imaging conditions across domains are indiscriminately collapsed into a single background category, leaving their variations unexplored.  (2) \textbf{Scene information loss:} rich full-scene observations are compressed into coarse binary labels, discarding scene content and target--context relations. (3) \textbf{Target supervision scarcity:} binary masks provide only pixel-wise supervision, and few-pixel targets therefore yield extremely sparse positive learning signals. 

Language supervision provides a natural means to recover the semantics omitted by binary masks. Accordingly, several methods have explored vision–language models for IRST detection by leveraging CLIP-based representations \cite{SAIST,textirstd}. However, CLIP's \emph{bag-of-words} \cite{bag} behavior favors isolated concepts over compositional relations, limiting fine-grained understanding of how few-pixel targets relate to their surrounding scenes. 

To address these challenges, we propose a new vision--language paradigm for Omni-IRST detection, termed ``understand before detect''. Specifically, it first instruction-tunes a generative vision--language model (VLM) on IRST understanding tasks and then transfers the learned visual representations to downstream dense prediction tasks. Different from task-specific supervised learning, this new paradigm employs an autoregressive objective to provide compositional supervision. As a result, it not only preserves rich scene-level information beyond binary classification, but also captures inherent domain-invariant semantics across heterogeneous infrared domains, providing reliable cues for detection. 

Despite this potential, the following challenges hinder this adoption for IRST detection: First, the few-pixel size and limited appearance cues of infrared small targets demand contextual reasoning, making existing object-centric instruction datasets and tasks unsuitable for direct adaptation. Second, the patchified and spatially coarse ViT features of VLMs are poorly suited to few-pixel localization, which requires fine-grained, multi-scale representations; adapting them through generic adapter-style modules is computationally costly and introduces redundant semantics.

To address these challenges, we propose JinSight (\textbf{J}oint Vision--Language Learning for \textbf{In}frared \textbf{Sight}), which establishes a holistic understanding of infrared scenes and targets, then transfers the acquired semantics to robust small target detection. In the first stage, visual instruction tuning employs language responses to learn IRST-aware representations across heterogeneous domains, producing instruction-tuned semantics for dense prediction. In the second stage, we introduce low-rank Latent Semantic Interaction (LSI), which efficiently exchanges instruction-tuned global semantics with multi-scale spatial features for precise small target detection. To address the lack of multimodal benchmarks for Omni-IRST detection, we construct OmniIRST-VL, the first large-scale Omni-IRST vision--language dataset, containing over 39k image--instruction--response annotations collected across diverse imaging platforms, spectral bands, and scene contexts. Its instructions are organized into two complementary groups: scene-oriented tasks characterize imaging conditions and scene backgrounds, whereas target-centric tasks reason about target presence, count, and location. Our main contributions are summarized as follows:
\begin{itemize}
	\item We propose a novel paradigm that reformulates Omni-IRST detection from pixel-wise binary prediction into holistic infrared understanding followed by detection, thereby preserving full-scene context, enriching supervision for infrared small targets, and learning transferable semantics across heterogeneous infrared domains.

	\item We propose JinSight, a two-stage vision--language framework: visual instruction tuning first learns IRST-aware representations across heterogeneous domains, and LSI then exchanges the instruction-tuned global semantics with fine-grained multi-scale features in a compact low-rank space, improving both accuracy and efficiency.
	
	\item We construct OmniIRST-VL, the first large-scale Omni-IRST vision--language dataset, comprising over 39k image--instruction--response annotations and six complementary instruction tasks. Comprehensive experiments demonstrate the effectiveness of JinSight, achieving an IoU improvement of over 14\% compared with state-of-the-art methods and providing a new benchmark for future research in Omni-IRST detection.
	
\end{itemize}

\begin{table*}[t]
	\centering
	\fontsize{8}{9}\selectfont
	\setlength{\tabcolsep}{0.7pt}
	\renewcommand{\arraystretch}{0.92}
	\def\tabyes{\checkmark}
	\def\tabyesbold{\ensuremath{\boldsymbol{\checkmark}}}
	\def\tabno{\ensuremath{\times}}
	
	\makebox[\textwidth][c]{%
		\begin{tabularx}{0.99\textwidth}{@{}>{\raggedright\arraybackslash}p{0.17\textwidth}>{\centering\arraybackslash}p{0.07\textwidth}|*{4}{>{\centering\arraybackslash}X}|*{2}{>{\centering\arraybackslash}X}|>{\centering\arraybackslash}X|*{3}{>{\centering\arraybackslash}X}|*{3}{>{\centering\arraybackslash}X}|>{\centering\arraybackslash}p{0.052\textwidth}>{\centering\arraybackslash}p{0.06\textwidth}@{}}
			\toprule
			\multirow{2}{*}{\textbf{Datasets}} & \multirow{2}{*}{\textbf{Venue}} & \multicolumn{4}{c|}{\textbf{Language Tasks}} & \multicolumn{2}{c|}{\textbf{Vision Tasks}} & \multirow{2}{*}{\makecell[c]{\textbf{IT}\\\textbf{Pairs}}} & \multicolumn{3}{c|}{\textbf{Imaging System}} & \multicolumn{3}{c|}{\textbf{Waveband}} & \multirow{2}{*}{\makecell[c]{\textbf{No. of}\\\textbf{Images}}} & \multirow{2}{*}{\makecell[c]{\textbf{Targets}\\\scriptsize$\leq10$ px}} \\[-1.5pt]
			\cmidrule(lr){3-6} \cmidrule(lr){7-8} \cmidrule(lr){10-12} \cmidrule(lr){13-15}
			& & \scriptsize P.VQA & \scriptsize Count. & \scriptsize Ground. & \scriptsize Caption & \scriptsize Seg. & \scriptsize Det. & & \scriptsize Land & \scriptsize Aerial & \scriptsize Space & \scriptsize NIR & \scriptsize SWIR & \scriptsize LWIR & \\[-1pt]
			\midrule
			NUST-SIRST & ICCV'19 & \tabno & \tabno & \tabno & \tabno & \tabyes & \tabno & -- & \tabyes & \tabno & \tabno & \tabno & \tabno & \tabyes & 10100 & 1973 \\
			SIRST-v2 & TGRS'23 & \tabno & \tabno & \tabno & \tabno & \tabyes & \tabyes & -- & \tabno & \tabyes & \tabno & \tabno & \tabno & \tabyes & 1024 & 84 \\
			NUDT-SIRST & TIP'22 & \tabno & \tabno & \tabno & \tabno & \tabyes & \tabno & -- & \tabyes & \tabyes & \tabno & \tabno & \tabno & \tabyes & 1327 & 473 \\
			IRSTD-1K & CVPR'22 & \tabno & \tabno & \tabno & \tabno & \tabyes & \tabno & -- & \tabyes & \tabno & \tabno & \tabno & \tabno & \tabyes & 1001 & 190 \\
			NUDT-SIRST-Sea & TGRS'23 & \tabno & \tabno & \tabno & \tabno & \tabyes & \tabno & -- & \tabno & \tabno & \tabyes & \tabyes & \tabno & \tabno & 5808 & 4844 \\
			MIRSTD & CVPR'25 & \tabno & \tabno & \tabno & \tabyes & \tabyes & \tabno & 2.7K & \tabyes & \tabyes & \tabno & \tabno & \tabno & \tabyes & 2731 & 721 \\
			FZDT & ICCV'25 & \tabno & \tabno & \tabno & \tabyes & \tabyes & \tabno & 2.7K & \tabyes & \tabyes & \tabno & \tabno & \tabno & \tabyes & 2755 & 742 \\
			\textbf{OmniIRST-VL} & \textbf{Ours} & \tabyesbold & \tabyesbold & \tabyesbold & \tabyesbold & \tabyesbold & \tabyesbold & \textbf{39.7k} & \tabyesbold & \tabyesbold & \tabyesbold & \tabyesbold & \tabyesbold & \tabyesbold & \textbf{11000} & \textbf{7421} \\
			\bottomrule
		\end{tabularx}
	}
	\caption{Comparison of different datasets across supported tasks, imaging systems, and wavebands. ``Targets $\leq10$ px'' counts 8-connected foreground components with an area of at most 10 pixels; -- denotes unavailable statistics. The NUST-SIRST count is based on 10,090 readable masks and excludes 10 corrupted masks. (IT: Image-Text; P.VQA: Physical VQA; Count.: Visual Counting; Ground.: Grounding; Seg.: Segmentation; Det.: Detection.)}
	\label{tab:dataset_comparison}
\end{table*}

\section{Related Work}

\subsection{Single-frame IRST detection}

\textbf{CNN-based detection method:} Recently, deep learning-based methods have been extensively explored for IRST detection, significantly advancing detection technologies \cite{infraredsurvey, DTUM,li2025probing}. Convolutional Neural Networks have emerged as the main approach, playing a pivotal role in this field \cite{IRSTD-1K, liu2024infrared, DaiACM}. To embed local contrast priors into a deep CNN framework, ALCNet \cite{aclnet} proposed an attentional local contrast network. Additionally, densely nested interactive feature fusion \cite{dnanet} is used to repetitively fuse and enhance features of different levels. Furthermore, UIU-Net \cite{uiu} constructed a backbone by embedding a tiny U-Net within a larger U-Net framework, thus facilitating multi-level and multi-scale representation learning.

\textbf{Transformer-based detection method: } 
However, CNN-based IRST detection methods are constrained by their limited receptive fields. To handle this, researchers adapt the transformer blocks to the CNN backbone. To extract and integrate local details and global information, MTU-Net \cite{MTUnet} applied a hybrid multilevel transformer-CNN encoder. SCTransNet \cite{sctransnet} leverages spatial-channel transformer blocks on skip connections to effectively model long-range information. To address the high computational complexity associated with transformers, Mamba-based state space models \cite{MiMISTD} \cite{SAMamba} have been integrated.

\subsection{Vision-Language models of IRST detection}

The application of vision--language paradigms \cite{soni2024earthdial} is severely constrained in the infrared domain, where both infrared imagery and aligned linguistic descriptions are scarce and homogeneous. This paucity of diverse, large-scale multimodal data fundamentally hinders the effective transfer of powerful foundation models \cite{zhang2024earthgpt}. Despite these challenges, several pioneering efforts have emerged to explore this integration. For instance, Text-IRSTD \cite{textirstd} and similar approaches \cite{leveraging, languagemoveirst} introduce textual descriptions as a supervisory signal to enhance the discriminative power of infrared features. Similarly, SAIST \cite{SAIST} leverages Contrastive Language-Image Pretraining (CLIP) \cite{clip} to guide the SAM, aiming to bridge the modality gap through cross-modal knowledge distillation and prompt-based learning. 

Existing vision--language methods primarily utilize pretrained VLMs (e.g., CLIP) to modulate visual features for improved small target extraction. However, without autoregressive generation, language serves primarily as a global semantic cue rather than a structured prediction target, leaving the complex relations between few-pixel targets and surrounding infrared scenes without explicit supervision. Moreover, they overlook the VLM's potential as a standalone perception backbone. In response, we introduce JinSight, a foundational framework that trains an IRST-specific VLM on the OmniIRST-VL dataset, a larger and diverse infrared vision--language dataset.

\begin{figure*}[!th]
	\centering
	\includegraphics[width=0.99\textwidth]{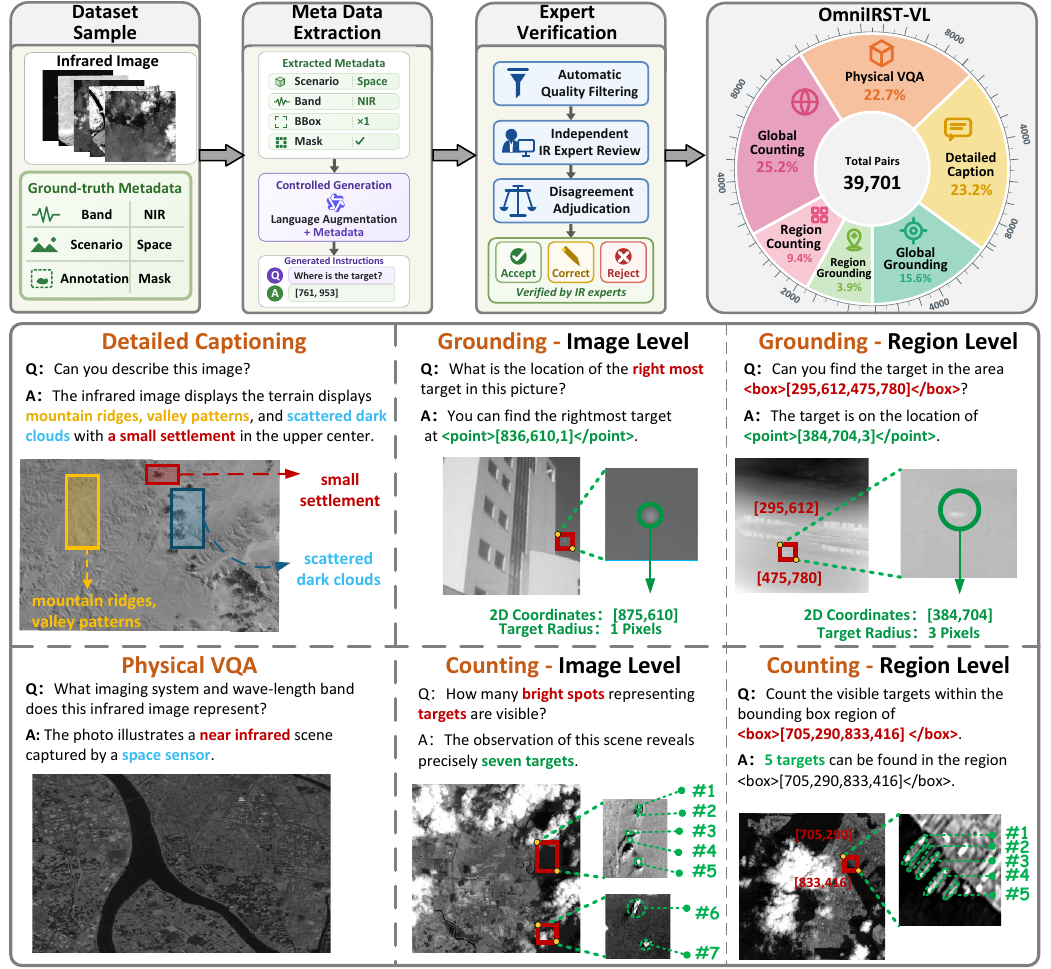}
	\caption{Overview of the construction, statistics, and task design of OmniIRST-VL. The top row illustrates the annotation pipeline, including metadata extraction, controlled instruction generation, automatic filtering, and expert verification, followed by the distribution of 39,701 image--instruction--response pairs across six tasks. The bottom rows present representative scene-oriented and target-centric examples spanning detailed captioning, physical VQA, counting, and grounding at image and region levels, demonstrating supervision from global scene understanding to fine-grained target reasoning.}\label{fig:datasets}
\end{figure*}

\begin{figure*}[!t]
	\centering
	\includegraphics[width=0.98\textwidth]{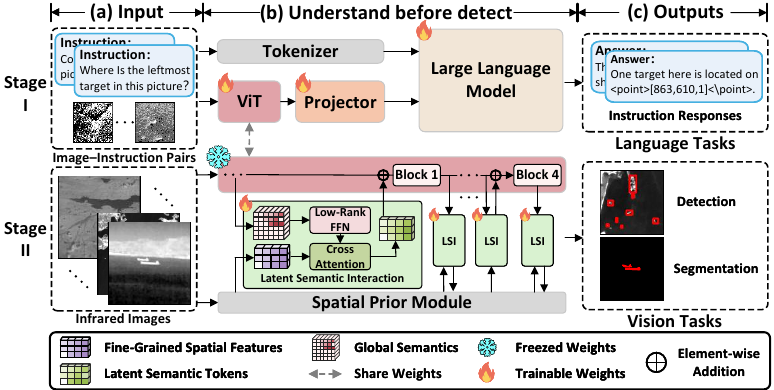}
	\caption{JinSight architecture. (a) Inputs for Stage I instruction tuning and Stage II dense prediction. (b) The ``understand before detect'' framework: Stage I learns an IRST-aware ViT via instruction supervision; Stage II uses LSI to fuse global semantics with fine-grained spatial features in a compact latent space. (c) Language and vision outputs.}
	\label{fig:Fig3}
\end{figure*}

\section{OmniIRST-VL Dataset}
\subsection{Data Collection and Annotation}

\subsubsection{Data Collection.}
To achieve broad coverage of infrared imaging domains, we adopt
WideIRSTD~\cite{wide} as the visual source. It contains
$11{,}000$ infrared images with $27{,}022$ annotated targets
collected under heterogeneous imaging conditions: image
resolutions range from $135\times96$ to $2700\times2900$, and
target areas range from $1$ to $1{,}094$ pixels. The images include
diverse environments such as urban areas, mountains, deserts,
maritime regions, and cloud clutter. As shown in
Table~\ref{tab:dataset_comparison}, OmniIRST-VL is the only dataset
spanning all three observation platforms and spectral bands and
contains the largest number of targets no larger than 10 pixels
($7{,}421$). These variations yield substantially
different background statistics and target--clutter
configurations, making WideIRSTD a suitable foundation for
omni-domain instruction construction.

\subsubsection{Data Annotation.}
As illustrated in Figure~\ref{fig:datasets}, we first extract
scenario and spectral-band attributes together with target masks and
bounding boxes from each image. We then perform metadata-conditioned
language augmentation to construct instruction--response pairs for
Detailed Captioning, Physical VQA, and image- and region-level Counting
and Grounding. The generated pairs undergo automatic quality filtering,
independent review by infrared experts, and disagreement adjudication;
each pair is accepted, corrected, or rejected to ensure semantic and
spatial reliability.

\subsection{Dataset Features and Statistics}
\label{sec:dataset_analysis}
\noindent\textbf{Dataset Splitting.}
Following the official WideIRSTD split, we construct
OmniIRST-Train from the training images and
OmniIRST-Benchmark from the test images.
The former is used for IRST-oriented instruction tuning, while the latter serves as
the evaluation benchmark, ensuring image-level separation between
training and evaluation.

\noindent\textbf{Omni-Domain Distribution.}
Table~\ref{tab:dataset_comparison} compares existing IRST datasets in
terms of supported tasks, imaging systems, and wavebands.
Unlike prior image-only or image-text datasets, OmniIRST-VL jointly
supports four language tasks and two vision tasks, while covering
land-, aerial-, and space-based imaging systems across NIR, SWIR, and
LWIR bands. With 39.7K image-text pairs from 11K images, it provides
broader task and domain coverage for Omni-IRST understanding.

\noindent\textbf{Rich Diversity and Spatial Granularity.}
Figure~\ref{fig:datasets} illustrates the diverse task types and
spatial granularities of OmniIRST-VL. The dataset covers six
instruction tasks with different semantic objectives and output
formats. Its instructions are organized into two complementary groups:
scene-oriented tasks characterize imaging conditions and scene
backgrounds, whereas target-centric tasks reason about target presence,
count, and location. The target-centric tasks operate at both image and
region levels, enabling models to understand targets over the entire
image or within a specified local area. Together, the two groups provide
complementary supervision from holistic background understanding to
fine-grained target reasoning.

\subsection{Evaluation Metrics}
We evaluate the performance across linguistic and visual dimensions. Linguistic capabilities are assessed via Accuracy for discriminative tasks and BLEU-1/4 \cite{bleu}, CIDEr \cite{cider}, and ROUGE-L \cite{rouge} for generative tasks, while perception performance is quantified through standard metrics: IoU, $P_d$, $F_a$, $AP_{50}$, Recall, and $F_1$. Detailed metric formulations are provided in the \textbf{supplementary material.}

\section{Method}
JinSight is a two-stage framework centered on a shared ViT
backbone. In Stage I, we fine-tune a generative VLM on
OmniIRST-VL with IRST-specific instructions, obtaining an
IRST-aware visual encoder that captures both scene context
and target-centric cues. The resulting model, JinSight-1B,
supports language-driven infrared understanding. In Stage II,
we transfer its adapted ViT to dense prediction, remove the
language projector and decoder, and introduce LSI modules
to exchange global semantic representations with
multi-scale spatial features. The resulting feature hierarchy
is fed into task-specific heads for detection and segmentation.
Thus, language acts as training-time representation supervision,
while dense prediction requires only infrared images at inference.

\subsection{IRST-Oriented Visual Instruction Tuning}

To instantiate the understanding stage, we instruction-tune the
pretrained VLM using image--instruction--response triples from
OmniIRST-VL. Starting from the generic ViT parameters $\theta_v$,
the autoregressive objective produces IRST-adapted parameters
$\theta_v^\star$ by requiring the visual representations to
support instruction-conditioned response generation.

Given an infrared image $I$, an instruction $q$, and its
response $y$, the model is optimized by
\begin{equation}
	\mathcal{L}_{\mathrm{inst}}
	=
	-\sum_{t=1}^{|y|}
	\log p(y_t\mid I,q,y_{<t}).
\end{equation}
Background-oriented tasks, including captioning and physical
VQA, supervise background and imaging characteristics,
whereas counting and grounding tasks require the encoder to
retain target-centric spatial evidence. Since response tokens
are conditioned on visual tokens, these complementary signals
jointly optimize the ViT to suppress clutter while preserving
weak target responses.

This objective benefits the visual encoder in two complementary ways: (1) \textbf{Language-Guided Semantic Alignment:} Language describes heterogeneous imaging conditions, scene backgrounds, and target cues through a shared vocabulary, encouraging visually diverse infrared observations to form domain-invariant semantic representations. (2) \textbf{Autoregressive Supervision:} Generating instruction-conditioned responses token by token requires the encoder to retain the spatial and semantic evidence needed to reason about target presence within the surrounding scene, thereby supervising information absent from binary masks.

\begin{table*}[t]
	\centering
	\footnotesize
	\renewcommand{\arraystretch}{0.90}
	\setlength{\tabcolsep}{2.20pt}
	\begin{tabular}{@{}l
			S[table-format=3.2,detect-weight=true,mode=text]
			*{15}{S[table-format=2.2,table-column-width=2.65em,
				detect-weight=true,mode=text]}@{}}
		\toprule
		\multirow{3}{*}{\textbf{Methods}}
		& \multicolumn{1}{c}{\multirow{3}{*}{\makecell[c]{\textbf{Params}\\\textbf{(M)}}}}
		& \multicolumn{6}{c}{\textbf{WideIRSTD}}
		& \multicolumn{9}{c}{\textbf{Cross-Dataset Generalization$^{*}$}} \\[-2pt]
		\cmidrule(lr){3-8} \cmidrule(lr){9-17}
			& & \multicolumn{3}{c}{\textbf{Segmentation}}
			& \multicolumn{3}{c}{\textbf{Detection}}
			& \multicolumn{3}{c}{\makecell[c]{\textbf{Synthetic}\\\textbf{Multi-Scene}}}
			& \multicolumn{3}{c}{\makecell[c]{\textbf{Real}\\\textbf{Variable-Resolution}}}
			& \multicolumn{3}{c}{\makecell[c]{\textbf{Real}\\\textbf{Complex-Background}}} \\[-2pt]
		\cmidrule(lr){3-5} \cmidrule(lr){6-8}
		\cmidrule(lr){9-11} \cmidrule(lr){12-14} \cmidrule(lr){15-17}
		& & {P$_d\uparrow$} & {IoU$\uparrow$} & {F$_a\downarrow$}
		& {AP$_{50}\uparrow$} & {Recall$\uparrow$} & {F$_1\uparrow$}
		& {P$_d\uparrow$} & {IoU$\uparrow$} & {F$_a\downarrow$}
		& {P$_d\uparrow$} & {IoU$\uparrow$} & {F$_a\downarrow$}
		& {P$_d\uparrow$} & {IoU$\uparrow$} & {F$_a\downarrow$} \\[-1pt]
		\midrule
		ACM & 0.40 & 51.92 & 32.55 & \bfseries 1.27 & 30.07 & 38.34 & 45.36 & 68.47 & 36.45 & 4.03 & 85.87 & 60.62 & 1.98 & 76.28 & 38.39 & 7.10 \\
		ResUnet & 0.91 & 47.88 & 31.39 & 1.58 & 29.57 & 38.91 & 47.52 & 75.13 & 48.17 & 7.49 & 90.33 & 67.49 & \underline{1.81} & 79.49 & \underline{45.07} & \underline{2.66} \\
		ALCNet & 0.52 & 56.95 & 32.26 & 2.79 & 31.98 & 40.26 & 42.51 & 79.37 & 45.06 & \underline{3.74} & 91.08 & 67.03 & 2.48 & 82.33 & 43.06 & 6.31 \\
		AMFU & 0.47 & 54.06 & 34.29 & 1.78 & 34.51 & 43.71 & 49.39 & 76.08 & 45.83 & 7.01 & \underline{93.42} & 66.12 & 2.90 & 81.09 & 38.79 & 8.66 \\
		U-Net & 34.53 & 58.21 & 33.89 & 2.51 & \underline{35.17} & 44.88 & 47.33 & 81.27 & 43.42 & 19.51 & 90.71 & 65.17 & 2.87 & 81.41 & 39.70 & 5.71 \\
		UIUNet & 50.54 & \underline{68.14} & \underline{37.46} & 2.34 & 33.58 & 49.70 & 45.73 & 74.71 & 45.01 & 4.66 & 90.71 & 68.10 & 2.62 & 71.79 & 38.48 & 2.85 \\
		DNANet & 4.70 & 57.16 & 34.45 & 2.33 & 34.63 & 46.11 & 47.94 & 81.27 & 46.49 & 13.69 & 92.19 & 66.96 & 2.93 & 65.38 & 31.19 & \bfseries 1.89 \\
		SCTransNet & 13.32 & 58.41 & 36.24 & 2.57 & 34.93 & 47.52 & \underline{50.00} & \underline{82.33} & \underline{52.34} & 5.74 & 92.94 & \underline{70.23} & \bfseries 1.52 & 80.45 & 38.47 & 3.29 \\
		NS-FPN & 4.17 & 63.38 & 29.48 & 5.71 & 24.65 & 50.78 & 35.72 & 71.53 & 42.48 & \bfseries 3.36 & 92.34 & 67.01 & 4.42 & 77.54 & 38.69 & 3.51 \\
		PConv-SD & 4.07 & 65.18 & 31.79 & 4.36 & 26.20 & \underline{52.11} & 46.72 & 70.05 & 41.77 & 5.30 & 88.51 & 64.75 & 3.32 & \underline{82.87} & 44.04 & 3.33 \\
		Text-IRSTD & 181.15 & 65.59 & 36.62 & 2.71 & 27.40 & 50.37 & 46.94 & 81.73 & 38.41 &  24.31 & 93.02 & 66.80 & 4.90 & 81.41 & 41.68 & 3.96 \\
		\textbf{JinSight} & 314.04 & \bfseries 73.75 & \bfseries 42.94 & \underline{1.51} & \bfseries 59.92 & \bfseries 53.68 & \bfseries 56.63 & \bfseries 86.24 & \bfseries 54.31 & 5.30 & \bfseries 94.79 & \bfseries 72.41 & 2.87 & \bfseries 83.72 & \bfseries 50.04 & 3.46 \\
		\bottomrule
	\end{tabular}

		\parbox{\textwidth}{\footnotesize\raggedright $^{*}$  Each test set is excluded from both Stage I instruction tuning and Stage II dense-prediction training of JinSight.}
	\caption{Quantitative comparison with state-of-the-art methods on the benchmark
		dataset and cross datasets. $F_a$ is reported in units of $10^{-5}$. The best and second-best results are highlighted in \textbf{bold} and \underline{underlined}, respectively.}
	\label{tab:sota_merged}
	\label{tab:Generalization}
\end{table*}

\subsection{Latent Semantic Interaction}

Although the instruction-tuned ViT provides discriminative
global semantics, its single-scale tokens lack the spatial
details required for localizing extremely small targets.
Direct high-dimensional fusion is inefficient because
background responses dominate infrared features. We therefore
introduce LSI, which exchanges information between ViT
tokens $V^i$ and multi-scale spatial
features $S^i$ in a compact latent space of dimension
$r\ll D$.

Specifically, each LSI block performs bidirectional interaction:
the spatial-to-semantic path transfers local details from $S^i$
to the ViT tokens, while the semantic-to-spatial path propagates
global IRST semantics in the reverse direction:
\begin{align}
	\widetilde V^i &=
	V^i + P_v^o
	\operatorname{Attn}(P_v^qV^i,P_s^kS^i,P_s^vS^i), \\
	\widetilde S^i &=
	S^i + P_s^o
	\operatorname{Attn}(P_s^qS^i,P_v^k\widetilde V^i,
	P_v^v\widetilde V^i).
\end{align}
All projections operate in the $r$-dimensional latent space.
A low-rank FFN subsequently refines the latent semantic tokens.
	Following ViT-Adapter \cite{vitadapter}, the resulting features
	are organized into a multi-scale pyramid and passed to a
	task-specific head \cite{upernet}.

\section{Experiments}
\subsection{Experimental settings}

In our experiments, we use the publicly available WideIRSTD dataset \cite{wide} and our proposed OmniIRST-VL benchmark to evaluate models’ capabilities in infrared small target understanding and detection. More detailed descriptions of the datasets and implementation details can be found in the \textbf{supplementary material}.

\subsection{Comparison to State-of-the-Art Methods}

\begin{table}[t]
	\centering
	\footnotesize
	\setlength{\tabcolsep}{2.2pt}
	\renewcommand{\arraystretch}{0.96}
	\begin{tabular}{@{}llcccc@{}}
			\toprule
			\textbf{Task} & \textbf{Metric}
			& \textbf{\makecell{InternVL\\2.5-1B}}
			& \textbf{\makecell{Qwen3\\VL-2B}}
			& \textbf{\makecell{LLaVA\\NeXT-7B}}
			& \textbf{Ours} \\
			\midrule
			Physical & Accuracy & \underline{32.35} & 19.20 & 21.10 & \textbf{49.80} \\
			\midrule
			\multirow{3}{*}{Counting}
			& Accuracy & 32.25 & 37.40 & \underline{43.40} & \textbf{53.20} \\
			& Global   & 35.56 & 35.08 & \underline{36.16} & \textbf{63.37} \\
			& Regional & 22.72 & \underline{44.08} & \textbf{64.27} & 23.88 \\
			\midrule
			\multirow{3}{*}{Grounding}
			& Accuracy & 30.66 & \underline{32.40} & 27.28 & \textbf{50.96} \\
			& Global   & 31.36 & \underline{33.98} & 27.98 & \textbf{51.96} \\
			& Regional & \underline{26.87} & 23.88 & 23.51 & \textbf{45.52} \\
			\midrule
			\multirow{3}{*}{Caption}
			& BLEU-1  & \underline{6.35} & 6.13 & 5.68 & \textbf{30.96} \\
			& CIDEr   & 0.09 & \underline{11.33} & 8.10 & \textbf{72.95} \\
			& ROUGE-L & \underline{10.33} & 9.54 & 8.80 & \textbf{27.70} \\
			\bottomrule
	\end{tabular}
	\caption{Quantitative comparison with large vision--language models on the OmniIRST-VL benchmark.}
	\label{tab:language}
\end{table}

\textbf{Result on OmniIRST-VL benchmark.} We conduct comprehensive comparisons with representative Large Vision--Language Models spanning different model scales, including InternVL2.5-1B~\cite{InternVL}, Qwen3-VL-2B~\cite{qwen3vl}, and LLaVA-NeXT-7B~\cite{llava-next}. Following the evaluation protocol of SkySenseGPT~\cite{skysense}, we evaluate these generic LVLMs in the zero-shot setting using standard metrics and compare them with JinSight-1B after Stage I instruction tuning.

\begin{figure}[t]
	\vspace{-0.1cm}
	\centering
	\includegraphics[width=\columnwidth]{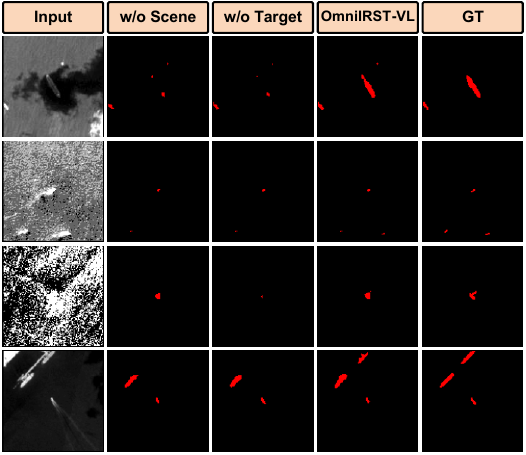}
\caption{Qualitative results of different instruction configurations in OmniIRST-VL. ``w/o scene'' removes scene-oriented instructions, ``w/o target'' removes target-centric instructions, and ``OmniIRST-VL'' uses all instructions.}\label{fig:visual_analysis}
\end{figure}
\noindent\textbf{Results on WideIRSTD.} We compare JinSight with 11 learning-based IRST detectors: ACM~\cite{DaiACM}, ALCNet~\cite{aclnet}, ResUNet~\cite{resunet}, AMFU~\cite{AMFU}, U-Net~\cite{unet}, DNA-Net~\cite{dnanet}, UIU-Net~\cite{uiu}, SCTransNet~\cite{sctransnet}, NS-FPN~\cite{seeing}, PConv-SD~\cite{pinwheel}, and Text-IRSTD~\cite{textirstd}. To ensure a fair comparison, we retrain all methods on the same training split under identical settings.

\noindent\textbf{Results on Cross-Dataset Generalization.} We conduct leave-one-dataset-out evaluation on NUDT-SIRST, NUAA-SIRST, and IRSTD-1K, representing three complementary domain shifts: synthetic multi-scene imagery, real variable-resolution observations, and real complex-background scenes, respectively. JinSight achieves the highest $P_d$ and IoU under all three shifts, demonstrating consistent transfer across heterogeneous domains.

\noindent\textbf{Quantitative Results:} As shown in Table~\ref{tab:sota_merged}, JinSight achieves the best performance across the primary metrics. We attribute this superiority to two key factors: 1) The ``understand before detect'' paradigm enables the visual backbone to retain scene context and target--context relations through generative visual instruction tuning, while LSI efficiently transfers these global semantics to multi-scale spatial features. This enhances the network's ability to distinguish few-pixel targets from heterogeneous background clutter. 2) OmniIRST-VL provides complementary scene-oriented and target-centric instructions across diverse infrared domains, enabling the backbone to learn domain-invariant scene and target semantics before detection. This produces more robust relational representations and improves cross-domain generalization.

\begin{table}[t]
	\centering
	
	\footnotesize
	\renewcommand{\arraystretch}{1.0}
	\begin{tabular*}{\columnwidth}{@{\extracolsep{\fill}}ccccc@{}}
		\toprule
			IVIT
		& LSI
		& $P_d \uparrow$
		& $IoU \uparrow$
		& $F_a \downarrow$ \\
		\midrule
		&
		& 70.28 & 40.72 & \textbf{1.21} \\

			\checkmark
		&
			& 72.34 & \underline{42.43} & 1.52 \\
		
			& \checkmark
		& 71.73 & 41.79 & 1.61 \\
		
		\checkmark
		& \checkmark
		& \textbf{73.75} & \textbf{42.94} & \underline{1.51} \\
		\bottomrule
	\end{tabular*}
		\caption{Ablation study of the proposed components. IVIT and LSI denote IRST-oriented visual instruction tuning and Latent Semantic Interaction, respectively.}
	\label{tab:component_ablation}
\end{table}

\begin{table}[t]
	\centering
	\footnotesize
	\renewcommand{\arraystretch}{1.0}
	\begin{tabular*}{\columnwidth}{@{\extracolsep{\fill}}lcccc@{}}
			\toprule
			\makecell{\textbf{Latent}\textbf{Dim.} $r$}
			& $IoU \uparrow$
			& $P_d \uparrow$
			& \makecell{Latency\\(ms) $\downarrow$}
			& \makecell{Memory\\(GB) $\downarrow$} \\
			\midrule
			w/o LSI  & 40.72 & 70.28 & 45.17 & 16.92 \\
			64       & 41.23 & 71.33 & \textbf{22.45} & \textbf{9.43} \\
			128      & \underline{41.79} & \textbf{71.73} & \underline{22.63} & \underline{11.63} \\
			256      & \textbf{42.16} & \underline{71.44} & 30.75 & 14.09 \\
			512      & 39.87 & 69.31 & 60.13 & 16.99 \\
		\bottomrule
	\end{tabular*}
	\caption{Effect of the LSI latent dimension $r$ on efficiency and detection performance.}
	\label{tab:rank_ablation}
\end{table}

\begin{table}[t]
	\centering
	\footnotesize
	\setlength{\tabcolsep}{6pt}
	\renewcommand{\arraystretch}{1.05}
	\begin{tabular}{@{}lccc@{}}
		\toprule
		\textbf{Instruction-Tuning Configuration}
		& $IoU \uparrow$
		& $P_d \uparrow$
		& $F_a \downarrow$ \\
		\midrule
		w/o OmniIRST-VL Instructions  & 40.72 & 70.28 & 1.21 \\
		w/o Scene-Oriented Instructions  & 41.63 & 71.00 & 1.23 \\
		w/o Target-Centric Instructions & 41.86 & 70.81 & 1.50 \\
		Full OmniIRST-VL Instructions                 & 42.43 & 72.34 & 1.52 \\
		\bottomrule
	\end{tabular}
	\caption{Ablation study of different instruction configurations in visual instruction tuning.}
	\label{tab:understanding_ablation}
\end{table}

\subsection{Ablation Study}
\textbf{Contribution of Components.} As shown in Table~\ref{tab:component_ablation}, adding IRST-oriented visual instruction tuning to the baseline yields relative improvements of 4.20\% in IoU and 2.93\% in $P_d$, demonstrating that language-guided IRST understanding provides beneficial semantic supervision for downstream detection. Adding LSI alone produces relative gains of 2.63\% in IoU and 2.06\% in $P_d$, showing that interacting global semantics with fine-grained spatial features facilitates dense target localization. Notably, using both components achieves the largest relative improvements, increasing IoU by 5.45\% and $P_d$ by 4.94\% over the baseline while reducing $F_a$ relative to either single-component setting. This indicates that instruction tuning provides IRST-aware semantic knowledge, while LSI effectively transfers these semantics into multi-scale spatial features; removing either semantic learning or semantic-spatial interaction limits the performance gains.

\noindent \textbf{The impact of the Latent Dimension of Latent Semantic Interaction.}
As shown in Table~\ref{tab:rank_ablation}, we investigate how the latent dimension $r$ affects semantic transfer and computational efficiency. Setting $r=256$ yields the optimal performance while retaining lower latency and memory consumption than the baseline without LSI. This trade-off arises because an overly compact latent space excessively compresses global semantics, whereas an excessively large one retains redundant semantics and weakens the efficiency of interaction.

\noindent\textbf{Impact of OmniIRST-VL Instruction Composition.} As shown in Table~\ref{tab:understanding_ablation} and Figure~\ref{fig:visual_analysis}, we evaluate the impact of instruction compositions in OmniIRST-VL. The full instruction set achieves the best IoU and Pd, demonstrating the complementary benefits of scene-oriented and target-centric supervision. Scene-oriented instructions enhance contextual reasoning, while target-centric instructions provide fine-grained localization cues. Together, they enable holistic IRST understanding prior to detection and improve representation transferability across heterogeneous domains.

\section{Conclusion}

We introduced ``understand before detect'', a new paradigm that reformulates omni-domain IRST detection from pixel-wise binary prediction into holistic infrared understanding followed by detection. Following this paradigm, we propose JinSight, which preserves full-scene context and target--context relations, enabling the visual backbone to learn transferable representations across heterogeneous infrared domains. A low-rank Latent Semantic Interaction exchanges its instruction-tuned global semantics with fine-grained multi-scale spatial features for dense prediction.  To support training and evaluation, we present OmniIRST-VL, an extended version of the WideIRSTD dataset enriched with image--instruction--response annotations. Experimental results demonstrate that our method  consistently outperforms state-of-the-art methods, achieving over 14\% improvement in IoU compared to existing methods.

\makeatletter
\gdef\showauthors@on{T}
\let\maketitle\arxivsavedmaketitle
\let\@maketitle\arxivsavedatmaketitle
\let\c@preprint@eqfn\relax
\let\thepreprint@eqfn\relax
\let\c@preprint@corrfn\relax
\let\thepreprint@corrfn\relax
\let\titlearea\relax
\let\actualheight\relax
\makeatother
\title{Understand Before Detect: Vision--Language Learning for Omni-Domain Infrared Small Target Detection}
\author{Supplementary Materials}
\affiliations{}

\maketitle
\vspace{-0.5in}

\noindent\textbf{Resource Availability.}
To support reproducibility and future research, the OmniIRST-VL dataset,
source code, training and evaluation configurations, and pretrained model
weights will be made publicly available upon publication.
\medskip

\setcounter{section}{6}
\setcounter{figure}{4}
\setcounter{table}{6}

\section{OmniIRST-VL Dataset}

\subsection{Challenges of Omni-Domain IRST Detection}
\label{sec:supp_task_challenges}

Omni-domain IRST (Omni-IRST) detection aims to localize infrared small
targets across heterogeneous imaging domains with a single unified detector.
Unlike conventional settings that are often evaluated within a specific
sensor, scene type, or acquisition platform, Omni-IRST detection must remain
robust to simultaneous shifts in scene context, background characteristics,
imaging conditions, and target appearance. WideIRSTD~\cite{wide}, which serves
as the visual source of OmniIRST-VL, exposes these challenges at scale: its
images range from $135\times96$ to $2700\times2900$ pixels, while target areas
range from only 1 to 1,094 pixels.

\begin{figure*}[!t]
    \centering
    \includegraphics[width=\textwidth]{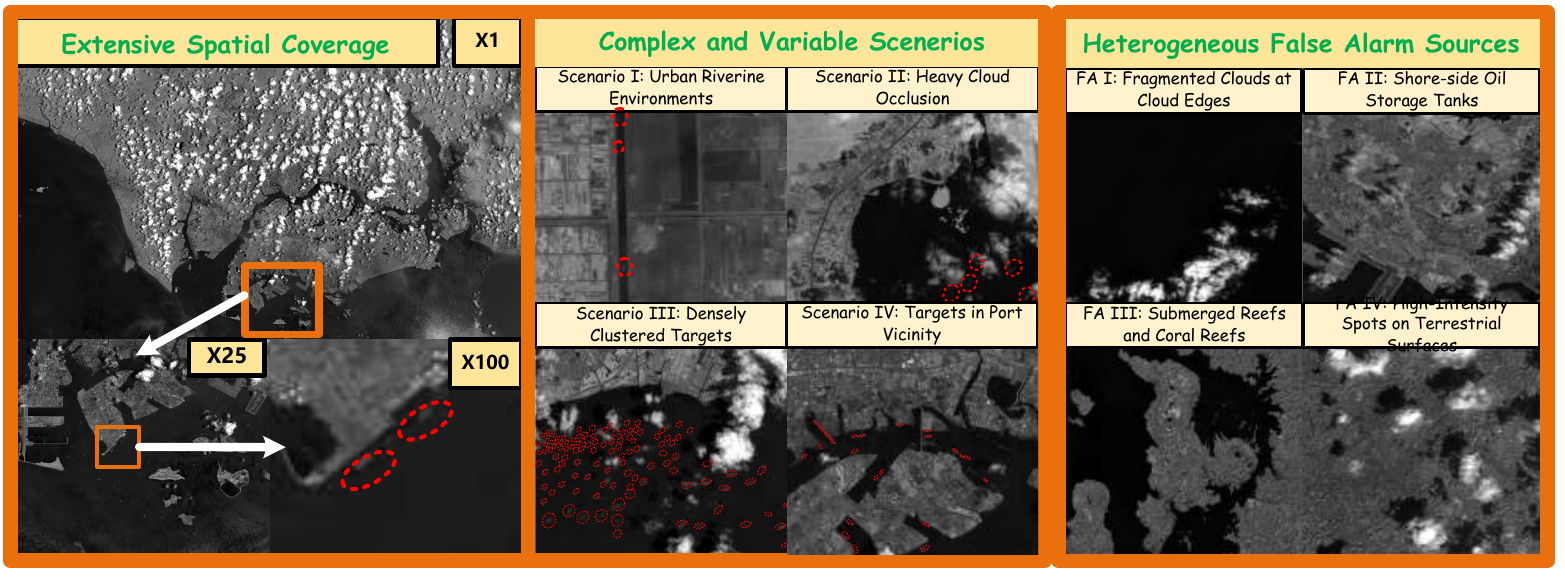}
    \caption{Representative challenges in WideIRSTD, the visual source of
    OmniIRST-VL. The enlarged
    regions illustrate how few-pixel targets can become nearly
    indistinguishable from surrounding clutter.}
    \label{fig:wideirstd}
\end{figure*}

As illustrated in Figure~\ref{fig:wideirstd}, three factors make unified
detection across these domains particularly difficult:

\begin{itemize}
    \setlength{\itemsep}{1pt}
    \setlength{\parsep}{0pt}
    \setlength{\topsep}{3pt}
    \item \textbf{Challenge I: Extensive Spatial Coverage.}
    Wide-domain IRST detection spans ground-based, airborne, and spaceborne
    observations with substantial variations in scene coverage and spatial
    resolution. In large-scale imagery, the target-to-background ratio becomes
    extremely small, and weak target responses can disappear during feature
    downsampling. This scale variation further compounds the domain shifts
    caused by heterogeneous imaging platforms.

    \item \textbf{Challenge II: Complex and Variable Scenarios.}
    Omni-IRST scenes cover structured urban regions, coastlines, cloud layers,
    and sea surfaces, producing large variations in background appearance and
    structural clutter. The same target may therefore appear under markedly
    different scene statistics, while structurally salient clutter can dominate
    the visual response. These scenario-dependent variations make it difficult
    for a unified detector to generalize to unseen imaging conditions.

    \item \textbf{Challenge III: Heterogeneous False Alarm Sources.}
    Cloud fragments, storage tanks, reefs, and high-intensity surface
    structures can exhibit local responses similar to true targets. At the
    same time, infrared small targets often occupy only a few pixels and thus
    provide very limited visual evidence. Consequently, distinguishing true
    targets from these heterogeneous hard negatives using local appearance
    alone is unreliable and can lead to both missed detections and false
    alarms.
\end{itemize}

Taken together, these task-level challenges expose the limitations of
conventional binary supervision discussed in the main paper. Variations in
spatial coverage, imaging platforms, scene content, clutter patterns, and
target-like distractors create pronounced domain heterogeneity. Binary masks
collapse this diversity into a single background category, discarding scene
information and target--context relations, while few-pixel targets yield
extremely sparse positive signals. These limitations motivate the ``\textbf{understand
before detect}'' paradigm adopted by JinSight. OmniIRST-VL supplements sparse
target masks with scene-oriented and target-centric language supervision,
enabling the visual backbone to learn domain-invariant scene semantics and
target--context relations before adapting them to precise dense prediction.

\subsection{Task Definitions}

We organize the annotations into four task families and six evaluation tracks.
Detailed captioning and physical VQA each define one track, whereas visual
grounding and visual counting are further divided into image-level and
region-level tracks. Together, these tasks provide complementary supervision
from holistic scene understanding to fine-grained target reasoning.

\begin{itemize}
    \setlength{\itemsep}{2pt}
    \setlength{\parsep}{0pt}
    \setlength{\topsep}{3pt}

    \item \textbf{Detailed Image Captioning.}
    Given an infrared image, the model generates a holistic description of the
    scene layout, background structures, prominent infrared responses, and
    target--context relations. In contrast to descriptions derived only from
    local target masks, these annotations preserve scene-level information that
    is useful for distinguishing true targets from structured clutter. This
    task contains $9{,}228$ image--instruction--response pairs.

    \item \textbf{Physical VQA.}
    This task queries imaging attributes rather than general scene semantics.
    The questions cover the spectral band, spatial resolution, and observation
    platform, using both direct and interrogative templates to keep their
    intent distinct from detailed captioning. We construct $9{,}000$ Physical
    VQA pairs from the WideIRSTD training split.

    \item \textbf{Image-Level and Region-Level Grounding.}
    Image-level grounding requires the model to search the entire image using a
    position-aware referring expression, whereas region-level grounding asks
    it to locate a target inside a specified query box. To accommodate extreme
    target-scale variation, targets with an area of at most 100 pixels use a
    point-based representation $[c_x,c_y,s]$, where $(c_x,c_y)$ is the target
    center and $s$ is a scale value derived from its spatial annotation. Larger
    targets use the bounding-box representation
    $[x_1,y_1,x_2,y_2]$. All coordinates are normalized to
    $[0,1000]$. The grounding set contains $7{,}760$ pairs, including $6{,}205$
    image-level and $1{,}555$ region-level queries. Its outputs comprise
    $5{,}794$ point-based annotations (74.7\%), $1{,}179$ box-based annotations
    (15.2\%), and 787 target-free negative samples (10.1\%). The negative
    samples explicitly evaluate whether a model hallucinates targets in empty
    scenes.

    \item \textbf{Image-Level and Region-Level Counting.}
    Image-level counting asks for the number of targets in the full image,
    while region-level counting restricts the query to a specified box. This
    design tests both numerical consistency and the ability to reject
    target-like background clutter. The counting set contains $13{,}713$ pairs,
    of which 73.0\% are image-level queries and 27.0\% are region-level queries.
\end{itemize}

The four task families contain $39{,}701$ image--instruction--response pairs in
total. During instruction tuning, the model learns from their native generative
responses, including textual descriptions, attribute answers, counts, and
spatial coordinates.
\subsection{OmniIRST-VL Benchmark Construction}

The held-out OmniIRST-Benchmark split is released in a standardized
four-option single-choice format for objective and reproducible evaluation.
Following recent remote-sensing VLM benchmarks that construct single-choice
questions from structured annotations and verify them manually
~\cite{lhrsbot,choice,geobenchvlm}, we derive every question and its unique
answer from the image metadata and target masks. Specifically, the metadata
provide the spectral band, spatial resolution, and observation platform for
Physical VQA, while the instance masks provide target presence, count, area,
center, and bounding-box coordinates for counting and grounding. Task-specific
templates render these records as questions, and the native answer is retained
as the sole correct option.

For each question, we then generate three task-consistent distractors in the
same output space as the correct answer. Physical-VQA distractors are sampled
from the valid attribute vocabulary and differ from the ground truth in one or
more queried attributes. Counting distractors are nearby nonnegative counts,
which prevent the question from being solved by coarse target-density
estimation. Grounding distractors use valid point or box syntax but correspond
to alternative locations that do not satisfy the referring expression or
region constraint. For target-free grounding questions, a designated
``no target'' response is the correct option and the three coordinate candidates
serve as hard negatives. We remove duplicated, out-of-range, or ambiguous
choices and require exactly one option to agree with the metadata and spatial
annotations. Finally, the four choices are randomly permuted and the correct
labels are balanced across `A', `B', `C', and `D'. Automatic consistency checks
are followed by manual review to verify question--image relevance, answer
uniqueness, and distractor plausibility. Detailed captioning remains an
open-ended generation task rather than being converted to multiple choice.

\section{Evaluation Metrics}

\subsection{OmniIRST-VL Evaluation Protocol}

The OmniIRST-VL benchmark evaluates the six tracks defined above: detailed
captioning, physical VQA, image-level grounding, region-level grounding,
image-level counting, and region-level counting. During instruction tuning,
all annotations retain their native generative response formats. For
standardized zero-shot evaluation, the discriminative tracks use the
four-option single-choice construction described above. A model is instructed
to return exactly one choice label from `A', `B', `C', and `D', and accuracy is
computed against the stored answer label. Detailed captioning remains an
open-ended generation task.

\begin{figure*}[!t]
	\centering
	\includegraphics[width=1.0\textwidth]{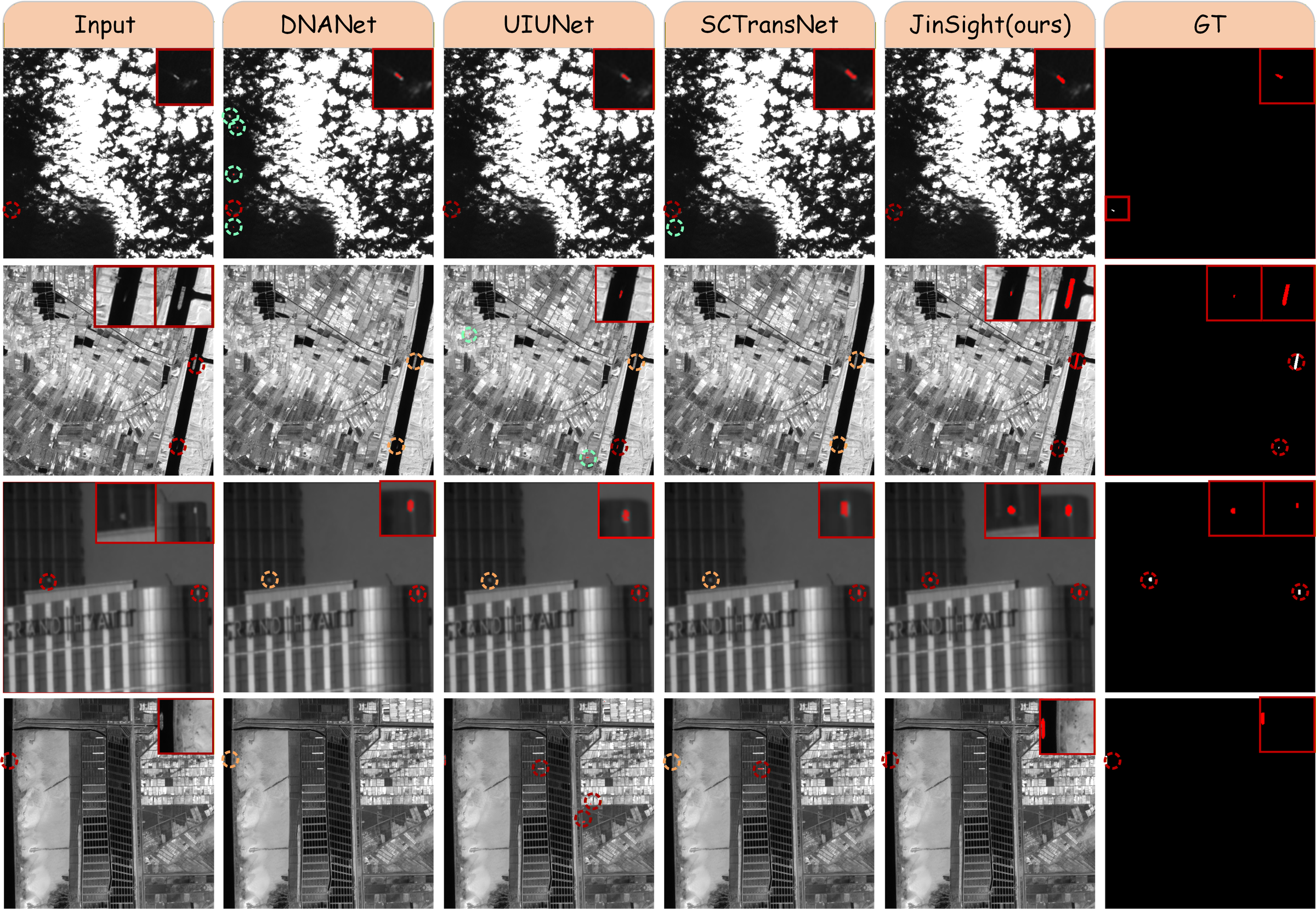}
	\caption{Qualitative comparison on representative challenging scenes.
		Red, orange, and green dashed circles indicate correct detections, missed
		targets, and false alarms, respectively. Enlarged regions highlight the
		responses of extremely small targets.}
	\label{fig:additional_qualitative_results}
\end{figure*}
\subsection{Metrics for Vision--Language Understanding}

\noindent\textbf{Accuracy.}
Physical VQA, grounding, and counting are evaluated by the average exact-match
accuracy of the predicted option labels:
\begin{equation}
    \mathrm{Acc}
    = \frac{1}{N}\sum_{i=1}^{N}
    \mathbb{I}\!\left(\hat{y}_i=y_i\right),
\end{equation}
where $N$ is the number of questions, $y_i$ and $\hat{y}_i$ are the ground-truth
and predicted option labels, respectively, and $\mathbb{I}(\cdot)$ is the
indicator function. We report the overall accuracy for each task family and
the separate image-level and region-level accuracies for grounding and
counting.

\noindent\textbf{BLEU.}
BLEU~\cite{bleu} measures the modified $n$-gram precision between a generated
caption and its reference while penalizing overly short outputs:
\begin{equation}
    \mathrm{BLEU}\text{-}N
    = \mathrm{BP}\exp\!\left(
      \sum_{n=1}^{N} w_n\log p_n
    \right),
\end{equation}
where $p_n$ is the modified $n$-gram precision, $w_n$ is its weight, and
$\mathrm{BP}$ is the brevity penalty.

\noindent\textbf{ROUGE-L.}
ROUGE-L~\cite{rouge} evaluates sequence-level agreement using the longest
common subsequence (LCS):
\begin{equation}
    F_{\mathrm{LCS}}
    = \frac{(1+\beta^2)R_{\mathrm{LCS}}P_{\mathrm{LCS}}}
    {R_{\mathrm{LCS}}+\beta^2P_{\mathrm{LCS}}},
\end{equation}
where $R_{\mathrm{LCS}}$ and $P_{\mathrm{LCS}}$ denote LCS-based recall and
precision, and $\beta$ controls their relative weighting.

\noindent\textbf{CIDEr.}
CIDEr~\cite{cider} measures consensus with the reference descriptions using
TF--IDF-weighted $n$-gram similarity:
\begin{equation}
    \mathrm{CIDEr}_n(c,S)
    = \frac{1}{M}\sum_{j=1}^{M}
    \frac{\boldsymbol{g}^{n}(c)\cdot\boldsymbol{g}^{n}(s_j)}
    {\lVert\boldsymbol{g}^{n}(c)\rVert
     \lVert\boldsymbol{g}^{n}(s_j)\rVert},
\end{equation}
where $c$ is the generated caption, $S=\{s_j\}_{j=1}^{M}$ is the reference set,
and $\boldsymbol{g}^{n}(\cdot)$ is the TF--IDF vector of $n$-grams.

\subsection{Metrics for Visual Perception}

\noindent\textbf{Intersection over Union (IoU).}
IoU evaluates pixel-level agreement between a predicted mask $P$ and the
ground-truth mask $G$:
\begin{equation}
    \mathrm{IoU}(P,G)=\frac{|P\cap G|}{|P\cup G|}.
\end{equation}

\noindent\textbf{Probability of Detection ($P_d$).}
$P_d$ measures target-level localization accuracy:
\begin{equation}
    P_d=\frac{N_c}{N_t},
\end{equation}
where $N_t$ is the number of ground-truth targets and $N_c$ is the number of
one-to-one matched detections. A prediction is considered correctly detected
when its centroid is less than three pixels from the centroid of the matched
ground-truth target.

\noindent\textbf{False Alarm Rate ($F_a$).}
$F_a$ is the ratio of unmatched positive pixels to all image pixels:
\begin{equation}
    F_a=\frac{P_f}{P_t},
\end{equation}
where $P_f$ is the number of pixels belonging to unmatched predicted
components and $P_t$ is the total number of evaluated pixels. Following IRST
convention, we report $F_a$ in units of $10^{-5}$.

\noindent\textbf{Average Precision at 0.5 IoU ($AP_{50}$).}
For cross-dataset detection evaluation, a prediction is a true positive when
its IoU with a one-to-one matched ground-truth instance is at least 0.5.
$AP_{50}$ is the area under the resulting precision--recall curve:
\begin{equation}
    AP_{50}=\int_{0}^{1}p_{50}(r)\,\mathrm{d}r,
\end{equation}
where $p_{50}(r)$ denotes precision as a function of recall at the 0.5 IoU
threshold.

\noindent\textbf{Recall and $F_1$ Score.}
Let $TP$, $FP$, and $FN$ denote the numbers of true-positive, false-positive,
and false-negative instances. Precision, recall, and their harmonic mean are
defined as
\begin{equation}
    \mathrm{Precision}=\frac{TP}{TP+FP},\qquad
    \mathrm{Recall}=\frac{TP}{TP+FN},
\end{equation}
\begin{equation}
    F_1=\frac{2\,\mathrm{Precision}\,\mathrm{Recall}}
    {\mathrm{Precision}+\mathrm{Recall}}.
\end{equation}

\begin{table*}[!t]
	\centering
	\footnotesize
	\setlength{\tabcolsep}{5pt}
	\renewcommand{\arraystretch}{1.20}
	\begin{tabularx}{\textwidth}{@{}p{0.16\textwidth}p{0.17\textwidth}>{\raggedright\arraybackslash}X@{}}
		\toprule
		\rowcolor{black!7}
		\textbf{Task} & \textbf{Scope or Variant} & \textbf{Prompts Sampled from the Instruction Data} \\
		\midrule
		\multirow{2}{*}{\textbf{Counting VQA}}
		& Image-level
		& \textbullet~Can you count how many targets are in this infrared scene?\newline
		\textbullet~How many small targets appear in this image?\newline
		\textbullet~What is the total number of visible targets?\newline
		\textbf{\ldots} \\
		\cmidrule(l){2-3}
		& Region-level
		& \textbullet~Count the targets within the bounding box \texttt{<box>[x1,y1,x2,y2]</box>}.\newline
		\textbullet~How many infrared targets can you detect in the area \texttt{<box>[x1,y1,x2,y2]</box>}?\newline
		\textbf{\ldots} \\
		\midrule
		\textbf{Detailed Captioning}
		& Image-level
		& \textbullet~What does this image show?\newline
		\textbullet~Can you describe what's shown here?\newline
		\textbullet~What objects or structures are visible?\newline
		\textbf{\ldots} \\
		\midrule
		\multirow{2}{*}{\textbf{Grounding VQA}}
		& Image-level
		& \textbullet~Identify the position of the target.\newline
		\textbullet~Can you locate the target in this infrared image?\newline
		\textbullet~Where is the target?\newline
		\textbf{\ldots} \\
		\cmidrule(l){2-3}
		& Region-level
		& \textbullet~Locate the target within the region \texttt{<box>[x1,y1,x2,y2]</box>}.\newline
		\textbullet~Within the region \texttt{<box>[x1,y1,x2,y2]</box>}, what is the position of the target?\newline
		\textbf{\ldots} \\
		\midrule
		\textbf{Physical VQA}
		& Image-level
		& \textbullet~Specify the physical properties, including wavelength band and imaging platform.\newline
		\textbullet~What platform and wavelength band were used to acquire this infrared image?\newline
		\textbullet~Classify this infrared image by its spectral band and acquisition platform.\newline
		\textbf{\ldots} \\
		\bottomrule
	\end{tabularx}
	\caption{Selected prompts sampled from the instruction data used to construct
		OmniIRST-VL. }
	\label{tab:vqa_examples}
\end{table*}
\section{Experiments}
This section provides the implementation details omitted from the main paper,
followed by supplementary experimental comparisons and an additional ablation
study.

\subsection{Implementation Details}

\noindent\textbf{Stage I: Visual Instruction Tuning.}

We initialize JinSight-1B from InternVL2.5-1B~\cite{InternVL} and train it on
OmniIRST-Train, with all OmniIRST-Benchmark images excluded. The six task tracks
are serialized using the native image--instruction--response chat template,
combined into a single instruction mixture, and shuffled during training.

We jointly optimize the visual backbone, language model, and multimodal
projector for three epochs. Training uses AdamW with an initial learning rate
of $4\times10^{-5}$, a global batch size of
128, weight decay of 0.01, a warm-up ratio of 0.03, and cosine learning-rate
decay. Training uses BF16 and gradient checkpointing. Images are processed at
$448\times448$ resolution using dynamic image tiling with at most six image
patches and one thumbnail per sample, and the maximum sequence length is 8,192
tokens. The language-modeling loss is computed only on the assistant response
tokens, while user-input and padding tokens are masked.

After Stage I, we retain the instruction-tuned ViT weights to initialize Stage
II and discard the language model and multimodal projector. For each
leave-one-dataset-out experiment, the held-out dataset is excluded from both
Stage-I and Stage-II training.

\noindent\textbf{Stage II: Dense Prediction Training.}

The dense-prediction stage is implemented with PyTorch 1.13.1 and
MMSegmentation 0.27.0 under CUDA 11.6 and is trained on four NVIDIA RTX 4090
GPUs. We initialize the ViT with the Stage-I instruction-tuned weights, set the
LSI latent dimension to $r=128$, and use UperNet~\cite{upernet} as the default
decoder. The model is optimized for 500 epochs with the SoftIoU loss and
AdamW~\cite{adamw}. The learning rate is linearly warmed up for 5,000
iterations to a peak value of $2\times10^{-4}$ and then decayed with a cosine
schedule. The total batch size is 32. All comparison methods are retrained on
the same data split under a consistent evaluation protocol.

\subsection{Experimental Comparisons}

In addition to the quantitative comparisons reported in the main paper,
Figure~\ref{fig:additional_qualitative_results} presents qualitative results
of representative methods on scenes with complex backgrounds and extremely
small targets. In the examples shown in
Figure~\ref{fig:additional_qualitative_results}, particularly in the fourth
row, JinSight detects the boundary target that is missed by several competing
methods and produces fewer visible false alarms.

\subsection{Additional Ablation Study}

\noindent\textbf{Effect of Different Decoders.}
JinSight supports different dense-prediction decoders as task-specific heads.
To isolate Stage-II architectural choices, both the latent-dimension ablation
reported in the main paper and the decoder ablation below use the pretrained
ViT initialization without Stage-I visual instruction tuning; all other
settings are fixed. In the decoder ablation, the LSI latent dimension is fixed
at $r=128$.
Table~\ref{tab:decoder_comparison} compares four representative decoders on
WideIRSTD. FPN yields the highest IoU and the lowest $F_a$, while SegFormer
achieves the highest $P_d$. UperNet ranks second across all three metrics and
is selected as the default decoder for its balanced performance.

\begin{table}[t!]
	\centering
	\small
	\setlength{\tabcolsep}{5pt}
	\renewcommand{\arraystretch}{1.15}
		\begin{tabular*}{\columnwidth}{@{\extracolsep{\fill}}lccc@{}}
		\toprule
		\textbf{Decoder}
		& $IoU \uparrow$
		& $P_d \uparrow$
		& $F_a \downarrow$ \\
		\midrule
		UperNet~\cite{upernet}     & \underline{41.79} & \underline{71.73} & \underline{1.61} \\
		DNANet~\cite{dnanet}       & 41.33 & 71.43 & 2.04 \\
		SegFormer~\cite{segformer} & 41.51 & \textbf{73.49} & 2.08 \\
		FPN~\cite{fpn}             & \textbf{42.74} & 71.17 & \textbf{1.43} \\
		\bottomrule
		\end{tabular*}
	\captionof{table}{Performance comparison of four representative decoders on
		WideIRSTD. The best and second-best results are highlighted in
		\textbf{bold} and \underline{underlined}, respectively.}
	\label{tab:decoder_comparison}
\end{table}

\section{Additional VQA Examples}
\label{sec:vqa_examples}

Table~\ref{tab:vqa_examples} presents selected prompts sampled directly from
the instruction data used to construct OmniIRST-VL, covering the four task
families and the image-level and region-level variants of grounding and
counting.

\FloatBarrier

\bibliography{references}

\begin{thebibliography}{46}
\providecommand{\natexlab}[1]{#1}

\bibitem[{Aibibu et~al.(2024)Aibibu, Lan, Zeng, Lu, and Gu}]{traffic}
Aibibu, T.; Lan, J.; Zeng, Y.; Lu, W.; and Gu, N. 2024.
\newblock Feature-enhanced attention and dual-gelan net (feadg-net) for uav
  infrared small object detection in traffic surveillance.
\newblock \emph{Drones}, 8(7): 304.

\bibitem[{An et~al.(2025)An, Sun, Gui, and He}]{choice}
An, X.; Sun, J.; Gui, Z.; and He, W. 2025.
\newblock Choice: benchmarking the remote sensing capabilities of large
  vision-language models.
\newblock In \emph{NeurIPS}, volume~38.

\bibitem[{Bai et~al.(2025)Bai, Cai, Chen, Chen, Chen, Cheng, Deng, Ding, Gao,
  Ge et~al.}]{qwen3vl}
Bai, S.; Cai, Y.; Chen, R.; Chen, K.; Chen, X.; Cheng, Z.; Deng, L.; Ding, W.;
  Gao, C.; Ge, C.; et~al. 2025.
\newblock Qwen3-vl technical report.
\newblock \emph{arXiv preprint arXiv:2511.21631}.

\bibitem[{Chen et~al.(2025)Chen, Ji, Peng, Zhu, Ye, and
  Sang}]{languagemoveirst}
Chen, S.; Ji, L.; Peng, S.; Zhu, S.; Ye, M.; and Sang, Y. 2025.
\newblock Language-driven motion prior knowledge learning for moving infrared
  small target detection.
\newblock \emph{IEEE Transactions on Geoscience and Remote Sensing}.

\bibitem[{Chen et~al.(2024{\natexlab{a}})Chen, Ye, Tan, Gong, Wu, Chu, Liu, Yu,
  and Ye}]{MiMISTD}
Chen, T.; Ye, Z.; Tan, Z.; Gong, T.; Wu, Y.; Chu, Q.; Liu, B.; Yu, N.; and Ye,
  J. 2024{\natexlab{a}}.
\newblock MiM-ISTD: Mamba-in-Mamba for Efficient Infrared Small-Target
  Detection.
\newblock \emph{IEEE Transactions on Geoscience and Remote Sensing}, 62: 1--13.

\bibitem[{Chen et~al.(2022)Chen, Duan, Wang, He, Lu, Dai, and
  Qiao}]{vitadapter}
Chen, Z.; Duan, Y.; Wang, W.; He, J.; Lu, T.; Dai, J.; and Qiao, Y. 2022.
\newblock Vision transformer adapter for dense predictions.
\newblock \emph{arXiv preprint arXiv:2205.08534}.

\bibitem[{Chen et~al.(2024{\natexlab{b}})Chen, Wu, Wang, Su, Chen, Xing, Zhong,
  Zhang, Zhu, Lu et~al.}]{InternVL}
Chen, Z.; Wu, J.; Wang, W.; Su, W.; Chen, G.; Xing, S.; Zhong, M.; Zhang, Q.;
  Zhu, X.; Lu, L.; et~al. 2024{\natexlab{b}}.
\newblock InternVL: Scaling up Vision Foundation Models and Aligning for
  Generic Visual-Linguistic Tasks.
\newblock In \emph{CVPR}.

\bibitem[{Chung, Lee, and Park(2023)}]{AMFU}
Chung, W.~Y.; Lee, I.~H.; and Park, C.~G. 2023.
\newblock Lightweight infrared small target detection network using full-scale
  skip connection U-Net.
\newblock \emph{IEEE Geoscience and Remote Sensing Letters}, 20: 1--5.

\bibitem[{Dai et~al.(2021{\natexlab{a}})Dai, Wu, Zhou, and Barnard}]{DaiACM}
Dai, Y.; Wu, Y.; Zhou, F.; and Barnard, K. 2021{\natexlab{a}}.
\newblock Asymmetric contextual modulation for infrared small target detection.
\newblock In \emph{Proceedings of the IEEE/CVF winter conference on
  applications of computer vision}, 950--959.

\bibitem[{Dai et~al.(2021{\natexlab{b}})Dai, Wu, Zhou, and Barnard}]{aclnet}
Dai, Y.; Wu, Y.; Zhou, F.; and Barnard, K. 2021{\natexlab{b}}.
\newblock Attentional local contrast networks for infrared small target
  detection.
\newblock \emph{IEEE Transactions on Geoscience and Remote Sensing}, 59(11):
  9813--9824.

\bibitem[{Danish et~al.(2025)Danish, Munir, Shah, Kuckreja, Khan, Fraccaro,
  Lacoste, and Khan}]{geobenchvlm}
Danish, M.; Munir, M.~A.; Shah, S. R.~A.; Kuckreja, K.; Khan, F.~S.; Fraccaro,
  P.; Lacoste, A.; and Khan, S. 2025.
\newblock Geobench-vlm: Benchmarking vision-language models for geospatial
  tasks.
\newblock In \emph{ICCV}, 7132--7142.

\bibitem[{Diakogiannis et~al.(2020)Diakogiannis, Waldner, Caccetta, and
  Wu}]{resunet}
Diakogiannis, F.~I.; Waldner, F.; Caccetta, P.; and Wu, C. 2020.
\newblock ResUNet-a: A deep learning framework for semantic segmentation of
  remotely sensed data.
\newblock \emph{ISPRS Journal of Photogrammetry and Remote Sensing}, 162:
  94--114.

\bibitem[{Guo et~al.(2024)Guo, Lao, Dang, Zhang, Yu, Ru, Zhong, Huang, Wu, Hu
  et~al.}]{skysense}
Guo, X.; Lao, J.; Dang, B.; Zhang, Y.; Yu, L.; Ru, L.; Zhong, L.; Huang, Z.;
  Wu, K.; Hu, D.; et~al. 2024.
\newblock Skysense: A multi-modal remote sensing foundation model towards
  universal interpretation for earth observation imagery.
\newblock In \emph{CVPR}, 27672--27683.

\bibitem[{Huang et~al.(2023)Huang, Li, Chen, Wang, Zhao, and Xu}]{antiuav}
Huang, B.; Li, J.; Chen, J.; Wang, G.; Zhao, J.; and Xu, T. 2023.
\newblock Anti-UAV410: A thermal infrared benchmark and customized scheme for
  tracking drones in the wild.
\newblock \emph{IEEE TPAMI}, 46(5): 2852--2865.

\bibitem[{Huang et~al.(2025)Huang, Zheng, Qiu, Liu, Bai, and Chen}]{textirstd}
Huang, F.; Zheng, S.; Qiu, Z.; Liu, H.; Bai, H.; and Chen, L. 2025.
\newblock Text-IRSTD: Leveraging Semantic Text to Promote Infrared Small Target
  Detection in Complex Scenes.
\newblock In \emph{ICCV}, 10635--10644.

\bibitem[{Li et~al.(2022)Li, Xiao, Wang, Wang, Lin, Li, An, and Guo}]{dnanet}
Li, B.; Xiao, C.; Wang, L.; Wang, Y.; Lin, Z.; Li, M.; An, W.; and Guo, Y.
  2022.
\newblock Dense nested attention network for infrared small target detection.
\newblock \emph{IEEE TIP}, 32: 1745--1758.

\bibitem[{Li et~al.(2024)Li, Ying, Li, Liu, Shi, and Li}]{wide}
Li, B.; Ying, X.; Li, R.; Liu, Y.; Shi, Y.; and Li, M. 2024.
\newblock The first competition on resource-limited infrared small target
  detection challenge: Methods and results.
\newblock \emph{arXiv preprint arXiv:2408.09615}.

\bibitem[{Li et~al.(2025)Li, An, Wang, Ying, Dai, Wang, Li, Guo, and
  Liu}]{li2025probing}
Li, R.; An, W.; Wang, Y.; Ying, X.; Dai, Y.; Wang, L.; Li, M.; Guo, Y.; and
  Liu, L. 2025.
\newblock Probing deep into temporal profile makes the infrared small target
  detector much better.
\newblock \emph{arXiv preprint arXiv:2506.12766}.

\bibitem[{Li et~al.(2023)Li, An, Xiao, Li, Wang, Li, and Guo}]{DTUM}
Li, R.; An, W.; Xiao, C.; Li, B.; Wang, Y.; Li, M.; and Guo, Y. 2023.
\newblock Direction-coded temporal U-shape module for multiframe infrared small
  target detection.
\newblock \emph{IEEE Transactions on Neural Networks and Learning Systems}.

\bibitem[{Lin(2004)}]{rouge}
Lin, C.-Y. 2004.
\newblock Rouge: A package for automatic evaluation of summaries.
\newblock In \emph{Text summarization branches out}, 74--81.

\bibitem[{Lin et~al.(2017)Lin, Doll{\'a}r, Girshick, He, Hariharan, and
  Belongie}]{fpn}
Lin, T.-Y.; Doll{\'a}r, P.; Girshick, R.; He, K.; Hariharan, B.; and Belongie,
  S. 2017.
\newblock Feature pyramid networks for object detection.
\newblock In \emph{CVPR}, 2117--2125.

\bibitem[{Liu et~al.(2024{\natexlab{a}})Liu, Li, Li, and Lee}]{llava-next}
Liu, H.; Li, C.; Li, Y.; and Lee, Y.~J. 2024{\natexlab{a}}.
\newblock Improved Baselines with Visual Instruction Tuning.
\newblock In \emph{CVPR}, 26286--26296.

\bibitem[{Liu et~al.(2024{\natexlab{b}})Liu, Zhang, Guo, and
  Ding}]{autonomous_driving}
Liu, P.; Zhang, Y.; Guo, G.; and Ding, J. 2024{\natexlab{b}}.
\newblock Enhanced detection and recognition of road objects in infrared
  imaging using multi-scale self-attention.
\newblock \emph{Sensors}, 24(16): 5404.

\bibitem[{Liu et~al.(2024{\natexlab{c}})Liu, Liu, Zheng, Wang, and
  Fu}]{liu2024infrared}
Liu, Q.; Liu, R.; Zheng, B.; Wang, H.; and Fu, Y. 2024{\natexlab{c}}.
\newblock Infrared small target detection with scale and location sensitivity.
\newblock In \emph{CVPR}, 17490--17499.

\bibitem[{Loshchilov and Hutter(2017)}]{adamw}
Loshchilov, I.; and Hutter, F. 2017.
\newblock Decoupled weight decay regularization.
\newblock \emph{arXiv preprint arXiv:1711.05101}.

\bibitem[{Muhtar et~al.(2024)Muhtar, Li, Gu, Zhang, and Xiao}]{lhrsbot}
Muhtar, D.; Li, Z.; Gu, F.; Zhang, X.; and Xiao, P. 2024.
\newblock Lhrs-bot: Empowering remote sensing with vgi-enhanced large
  multimodal language model.
\newblock In \emph{ECCV}, 440--457.

\bibitem[{Papineni et~al.(2002)Papineni, Roukos, Ward, and Zhu}]{bleu}
Papineni, K.; Roukos, S.; Ward, T.; and Zhu, W.-J. 2002.
\newblock Bleu: a method for automatic evaluation of machine translation.
\newblock In \emph{ACL}, 311--318.

\bibitem[{Radford et~al.(2021)Radford, Kim, Hallacy, Ramesh, Goh, Agarwal,
  Sastry, Askell, Mishkin, Clark et~al.}]{clip}
Radford, A.; Kim, J.~W.; Hallacy, C.; Ramesh, A.; Goh, G.; Agarwal, S.; Sastry,
  G.; Askell, A.; Mishkin, P.; Clark, J.; et~al. 2021.
\newblock Learning Transferable Visual Models From Natural Language
  Supervision.
\newblock In \emph{ICML}, 8748--8763.

\bibitem[{Ronneberger, Fischer, and Brox(2015)}]{unet}
Ronneberger, O.; Fischer, P.; and Brox, T. 2015.
\newblock U-net: Convolutional networks for biomedical image segmentation.
\newblock In \emph{International Conference on Medical image computing and
  computer-assisted intervention}, 234--241.

\bibitem[{Sagar~Soni(2025)}]{soni2024earthdial}
Sagar~Soni, H. D. M. F. M. A. M. M. S. D. P. F. C. W. L. J. K. S. K. F.~K.,
  Akshay~Dudhane. 2025.
\newblock EarthDial: Turning Multi-sensory Earth Observations to Interactive
  Dialogues.
\newblock \emph{ArXiv}.

\bibitem[{Shi et~al.(2025)Shi, He, Hui, Li, Yang, Cheng, and Dai}]{auxdet}
Shi, Y.; He, R.; Hui, L.; Li, X.; Yang, J.; Cheng, M.-M.; and Dai, Y. 2025.
\newblock AuxDet: Auxiliary Metadata Matters for Omni-Domain Infrared Small
  Target Detection.
\newblock \emph{arXiv e-prints}, arXiv--2505.

\bibitem[{Singh and Singh(2025)}]{leveraging}
Singh, P.; and Singh, P. 2025.
\newblock Leveraging Language Prior for Infrared Small Target Detection.
\newblock \emph{arXiv preprint arXiv:2507.13113}.

\bibitem[{Vedantam, Lawrence~Zitnick, and Parikh(2015)}]{cider}
Vedantam, R.; Lawrence~Zitnick, C.; and Parikh, D. 2015.
\newblock Cider: Consensus-based image description evaluation.
\newblock In \emph{CVPR}, 4566--4575.

\bibitem[{Wu et~al.(2023)Wu, Li, Luo, Wang, Xiao, Liu, Yang, An, and
  Guo}]{MTUnet}
Wu, T.; Li, B.; Luo, Y.; Wang, Y.; Xiao, C.; Liu, T.; Yang, J.; An, W.; and
  Guo, Y. 2023.
\newblock MTU-Net: Multilevel TransUNet for Space-Based Infrared Tiny Ship
  Detection.
\newblock \emph{IEEE Transactions on Geoscience and Remote Sensing}, 61: 1--15.

\bibitem[{Wu, Hong, and Chanussot(2022)}]{uiu}
Wu, X.; Hong, D.; and Chanussot, J. 2022.
\newblock UIU-Net: U-Net in U-Net for infrared small object detection.
\newblock \emph{IEEE TIP}, 32: 364--376.

\bibitem[{Xiao et~al.(2018)Xiao, Liu, Zhou, Jiang, and Sun}]{upernet}
Xiao, T.; Liu, Y.; Zhou, B.; Jiang, Y.; and Sun, J. 2018.
\newblock Unified perceptual parsing for scene understanding.
\newblock In \emph{ECCV}, 418--434.

\bibitem[{Xie et~al.(2021)Xie, Wang, Yu, Anandkumar, Alvarez, and
  Luo}]{segformer}
Xie, E.; Wang, W.; Yu, Z.; Anandkumar, A.; Alvarez, J.~M.; and Luo, P. 2021.
\newblock SegFormer: Simple and efficient design for semantic segmentation with
  transformers.
\newblock \emph{NeurIPS}, 34: 12077--12090.

\bibitem[{Xu et~al.(2025)Xu, Zheng, Wang, Zhang, Ren, Xu, and Xu}]{SAMamba}
Xu, W.; Zheng, S.; Wang, C.; Zhang, Z.; Ren, C.; Xu, R.; and Xu, S. 2025.
\newblock SAMamba: Adaptive state space modeling with hierarchical vision for
  infrared small target detection.
\newblock \emph{Information Fusion}, 124: 103338.

\bibitem[{Yang et~al.(2025)Yang, Liu, Wu, Su, Hai, and Huang}]{pinwheel}
Yang, J.; Liu, S.; Wu, J.; Su, X.; Hai, N.; and Huang, X. 2025.
\newblock Pinwheel-shaped convolution and scale-based dynamic loss for infrared
  small target detection.
\newblock In \emph{Proceedings of the AAAI Conference on Artificial
  Intelligence}, volume~39, 9202--9210.

\bibitem[{Yuan et~al.(2026)Yuan, Meng, Xi, Zhao, Zhao, Dai, and Wei}]{seeing}
Yuan, M.; Meng, D.; Xi, Z.; Zhao, T.; Zhao, S.; Dai, Y.; and Wei, X. 2026.
\newblock Seeing Through the Noise: Improving Infrared Small Target Detection
  and Segmentation from Noise Suppression Perspective.
\newblock In \emph{Proceedings of the IEEE/CVF Conference on Computer Vision
  and Pattern Recognition}, 27783--27792.

\bibitem[{Yuan et~al.(2024)Yuan, Qin, Yan, Akhtar, and Mian}]{sctransnet}
Yuan, S.; Qin, H.; Yan, X.; Akhtar, N.; and Mian, A. 2024.
\newblock Sctransnet: Spatial-channel cross transformer network for infrared
  small target detection.
\newblock \emph{IEEE Transactions on Geoscience and Remote Sensing}, 62: 1--15.

\bibitem[{Yuksekgonul et~al.(2023)Yuksekgonul, Bianchi, Kalluri, Jurafsky, and
  Zou}]{bag}
Yuksekgonul, M.; Bianchi, F.; Kalluri, P.; Jurafsky, D.; and Zou, J. 2023.
\newblock When and why Vision-Language Models behave like Bags-of-Words, and
  what to do about it?
\newblock In \emph{International Conference on Learning Representations}.

\bibitem[{Zhang et~al.(2025)Zhang, Li, Gao, Guo, Gao, and Zhang}]{SAIST}
Zhang, M.; Li, X.; Gao, F.; Guo, J.; Gao, X.; and Zhang, J. 2025.
\newblock SAIST: Segment any infrared small target model guided by contrastive
  language-image pretraining.
\newblock In \emph{CVPR}, 9549--9558.

\bibitem[{Zhang et~al.(2022)Zhang, Zhang, Yang, Bai, Zhang, and Guo}]{IRSTD-1K}
Zhang, M.; Zhang, R.; Yang, Y.; Bai, H.; Zhang, J.; and Guo, J. 2022.
\newblock ISNet: Shape matters for infrared small target detection.
\newblock In \emph{CVPR}, 877--886.

\bibitem[{Zhang et~al.(2024)Zhang, Cai, Zhang, Zhuang, and
  Mao}]{zhang2024earthgpt}
Zhang, W.; Cai, M.; Zhang, T.; Zhuang, Y.; and Mao, X. 2024.
\newblock EarthGPT: A Universal Multi-modal Large Language Model for
  Multi-sensor Image Comprehension in Remote Sensing Domain.
\newblock \emph{IEEE Transactions on Geoscience and Remote Sensing}, 62: 1--27.

\bibitem[{Zhao et~al.(2022)Zhao, Li, Li, Hu, Ma, and Tao}]{infraredsurvey}
Zhao, M.; Li, W.; Li, L.; Hu, J.; Ma, P.; and Tao, R. 2022.
\newblock Single-Frame Infrared Small-Target Detection: A survey.
\newblock \emph{IEEE Geoscience and Remote Sensing Magazine}, 10(2): 87--119.

\end{thebibliography}
\end{document}